# RSI-Net: Two-Stream Deep Neural Network for Remote Sensing Images-based Semantic Segmentation

Shuang He[1,2], Xia Lu[1,2], Jason Gu[3], Senior Member, IEEE, Haitong Tang[1,2], Qin Yu[4], Kaiyue Liu[1,2], Haozhou Ding[5], Chunqi Chang[6,7] and Nizhuan Wang[1,2,4,8]

[1]Jiangsu Key laboratory of Marine Bioresources and Environment/Jiangsu Key laboratory of Marine Biotechnology/Co-Innovation Center of Jiangsu Marine Bio-Industry Technology, Jiangsu Ocean University, Lianyungang, 222005, China
[2]School of Geomatics and Marine Information, Jiangsu Ocean University, Lianyungang, 222005, China
[3]Department of Electrical and Computer Engineering, Dalhousie University, Halifax, NS B3H 4R2, Canada
[4]School of Computer Engineering, Jiangsu Ocean University, Lianyungang, 222005, China
[5]School of Computer and Software, Nanjing University of Information Science and Technology, Nanjing, 211800, China
[6]School of Biomedical Engineering, Health Science Center, Shenzhen University, Shenzhen, 518060, China
[7]Peng Cheng Laboratory, Shenzhen, 518055, China
[8]School of Biomedical Engineering, ShanghaiTech University, Shanghai, 201210, China.

Corresponding authors: Nizhuan Wang (wangnizhuan1120@gmail.com), Xia Lu (2008000070@jou.edu.cn), Chunqi Chang (cqchang@szu.edu.cn).

This study was supported by the National Natural Science Foundation of China (No. 41506106, 61701318, 61971289), Project of "Six Talent Peaks" of Jiangsu Province (No. SWYY-017), Shenzhen Fundamental Research Project (No. JCYJ20170412111316339), Shenzhen-Hong Kong Institute of Brain Science-Shenzhen Fundamental Research Institutions, Project of Huaguoshan Mountain Talent Plan - Doctors for Innovation and Entrepreneurship, Jiangsu University Superior Discipline Construction Project Funding Project (PAPD)), and Jiangsu Province Graduate Research and Practice Innovation Program Project (No. KYCX2021-027).

**ABSTRACT** For semantic segmentation of remote sensing images (RSI), trade-off between representation power and location accuracy is quite important. How to get the trade-off effectively is an open question, where current approaches of utilizing very deep models result in complex models with large memory consumption. In contrast to previous work that utilizes dilated convolutions or deep models, we propose a novel two-stream deep neural network for semantic segmentation of RSI (RSI-Net) to obtain improved performance through modeling and propagating spatial contextual structure effectively and a decoding scheme with image-level and graph-level combination. The first component explicitly models correlations between adjacent land covers and conduct flexible convolution on arbitrarily irregular image regions by using graph convolutional network, while densely connected atrous convolution network (DenseAtrousCNet) with multi-scale atrous convolution can expand the receptive fields and obtain image global information. Extensive experiments are implemented on the Vaihingen, Potsdam and Gaofen RSI datasets, where the comparison results demonstrate the superior performance of RSI-Net in terms of overall accuracy (91.83%, 93.31% and 93.67% on three datasets, respectively), F1 score (90.3%, 91.49% and 89.35% on three datasets, respectively) and kappa coefficient (89.46%, 90.46% and 90.37% on three datasets, respectively) when compared with six state-of-the-art RSI semantic segmentation methods.

## I. INTRODUCTION

Semantic segmentation of high-resolution remote sensing image (RSI) can help to extract the land use and land cover (LULC) information [1-5], and has diverse applications, e.g., monitoring [6, 7], planning [8, 9], and detection [10, 11]. However, RSI due to the topography with different characteristics, different terrain and different scales are often quite distinct, and the image boundary details of various shapes also make the image segmentation task very challenging.

Conventional methods employ handcrafted features for machine learning [12-15]. Recently deep convolutional neural networks (DCNNs) [16, 17] have achieved great success in the semantic segmentation of natural scene by learning feature representations and classifier parameters simultaneously. DCNN-based methods obtain high-level features of images but may lose resolution as well and therefore fail to effectively utilize context information, resulting in rough segmentation and discontinuous boundaries. Advanced neural network models such as fully convolutional network (FCN) [18], SegNet [19], U-Net [20] and DeepLab [21] can effectively retain location information while extracting high-level features. The capability of restoring pixel categories makes FCN suitable for image semantic segmentation but lacking spatial consistency since relationships between pixels are not fully considered, while SegNet and U-Net are computationally intensive. In DeepLab [21-23], a fully connected conditional random field (CRF) layer is added to the end of the FCN to further exploit relationships among pixels, and the whole network is trained in Zheng *et al.* [24] in an end-to-end way.

Traditional RSI segmentation models are designed for Euclidean data and often ignore the inherent correlation between adjacent land covers, while the recently developed graph convolutional networks (GCNs) work directly on structural graphs [25, 26] and can be naturally applied to model spatial relations in RSIs, as demonstrated in Mou *et al.* [27] for hyperspectral image segmentation. If each pixel is treated as a graph node, then the computation of GCN is highly demanded due to the enormous number of pixels in high-resolution RSIs, therefore in [28] superpixels instead of pixels are utilized as the nodes to greatly reduce the size of the GCN. Though superpixel-based GCN can model various spatial structures of land cover, it cannot generate subtle individual features for each pixel. In contrast, CNN can learn locally spectral and spatial features at the pixel level, but its receptive field is usually limited to a small square window.

Thus, in this article, a novel dual-stream deep neural network integrating advantages of both GCN and CNN is proposed for semantic segmentation of RSIs, where GCN is utilized to work on graph-level features which have clearer semantic references than image-level features, and CNN with expanded receptive field due to atrous convolution is also included to capture pixel-level multi-scale contextual information, with effectiveness demonstrated on three pubic datasets. More specifically, the main contributions of this paper are summarized as follows.

1) A novel end-to-end two-stream deep neural network integrating CNN and GCN, named RSI-Net is proposed to process semantic information at both image and graph levels for improved semantic segmentation of RSIs. It includes a two-stream (GCN and CNN) encoder and a fused decoder.

2) In the CNN stream, multi-scale atrous convolution is utilized to expand the receptive fields for improved capability of learning local information and correlations, which help to capture more boundary related semantic information from RSIs.

3) The novel fused decoder incorporates both high and low level as well as graph-level features to achieve improved semantic segmentation performance.

4) Experimental results on three public RSI datasets demonstrate the superior performance of the proposed RSI-Net as compared to a number of state-of-the-art counterparts.

The rest organization of this article is given as follow. First, the Section II describes the related works and background. Then, details of the proposed model and embedding modules are provided in Section III. Next, the Section IV-VI depict the experimental design, results analysis and discussions. Finally, Section VII concludes our main work and feature work.

## II. RELATED WORKS

### A. Fully Convolutional Network (FCN) Based Structure for Semantic Segmentation

FCN is a fully convolutional network compatible with arbitrary size images and performs semantic segmentation in a fully supervised learning manner [18]. Skip-layers structure is utilized in FCN to achieve the complement of shallow detail information. Many of the state-of-the-art semantic segmentation methods are based on FCNs. Instead of following the jump connection structure of the typical FCN, the SegNet [19] is an encoder-decoder network with an innovation in upsampling, which recovers detailed information by drawing on the maximum pooling index in the decoding process, while U-Net [20] utilizes more shallow features to supplement detailed information. Since the aforementioned semantic segmentation models do not make good use of contextual information and thus lead to unsatisfactory segmentation, a novel model named PSPNet with a Pyramid Pooling Module (PPM) is proposed in Zhao *et al.* [29] to establish connections between contexts for improved feature representation, but it has the problem of losing pixels when increasing the field of sensation by pooling operations, which may be alleviated by atrous convolution [30] that can obtain a different perception of the field without changing the amount of computation as shown in Figure 1. However, atrous convolution suffers from the severe problem of losing local information, especially for large null rate and small targets. The hybrid atrous convolution [31] and Deeplab series [21-23] have been developed to effectively alleviate the problem of losing local information.

### B. Graph Convolutional Network (GCN)

A graph is a nonlinear structure used in a non-Euclidean space to represent connections. An undirected graph [32] is defined as $\mathcal{G} = (\upsilon, \varepsilon, \mathbf{A})$, where $\upsilon$ denotes the vertex set with $|\upsilon| = N$, $\varepsilon$ denotes the edge set of the graph, and $\mathbf{A} \in R^{N \times N}$ indicates the adjacent matrix, each entry $a_{ij}$ represents edge's weight between vertex $i$ and vertex $j$, respectively.

We denote the diagonal degree matrix of $\mathbf{A}$ as $\mathbf{D}$, whose entry $d_{ii} = \sum_{j=1}^{N} a_{ij}$ [32]. Then, the Laplacian matrix of graph $\mathcal{G}$ can be defined as

$$\boldsymbol{L} = \mathbf{D} - \mathbf{A}. \qquad (1)$$

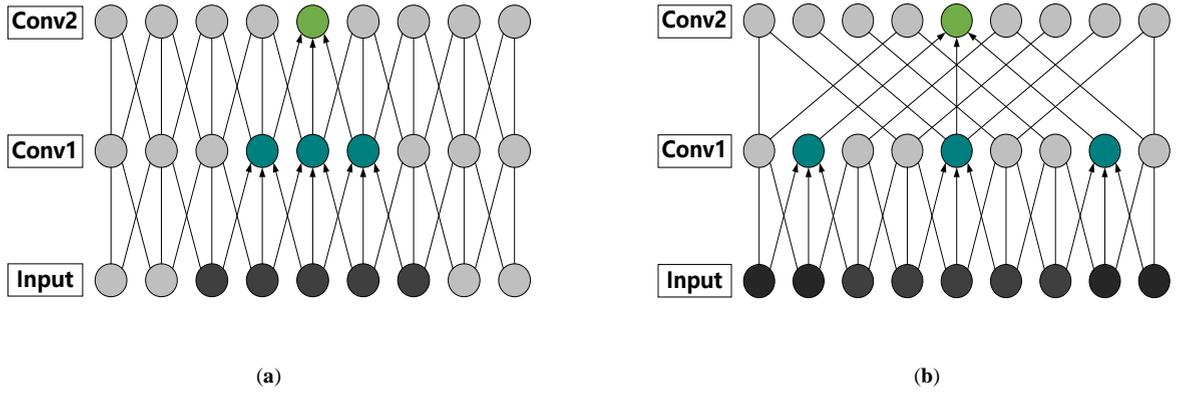

Fig. 1. Schematic diagram of reception fields of normal convolution and atrous convolution. (a) Description of normal convolution; (b) Description of atrous convolution.

To enhance the generalization ability of the graph, the symmetric normalized Laplacian Matrix is defined as

$$L_{sym} = I - D^{-\frac{1}{2}}AD^{-\frac{1}{2}} \qquad (2)$$

where $I$ is an identity matrix. Shuman *et al.* [33-35] extended the Fourier signals transformation specified on graphs and created the graph-structured data transition operation. Graph convolution of a graph-structured signal $x \in R^N$ with a filter $g_\theta$, is defined as

$$g_\theta \otimes x = U g_\theta(\Lambda) U^T x \qquad (3)$$

where $\Lambda$ represents the diagonal matrix of the eigenvalues of the normalized graph Laplacian $L_{sym}$, with $A$ and $D$ as the adjacency matrix and degree matrix of $x$, respectively; $U$ represents the eigenvectors of $L$; and $g_\theta(\Lambda)$ is the Fourier transform [33] of $g_\theta$, which can be calculated as

$$g_\theta(\Lambda) = \int_R g_\theta e^{-2\pi i \Lambda t} d_t. \qquad (4)$$

In addition, $U^T x$ denotes the graph Fourier transform of $x$. $g_\theta(\Lambda)$ can be framed as a function of the eigenvalues of $\Lambda$.

However, it can be noted that Eq. (3) requires explicitly calculating the Laplacian eigenvectors, which are not computationally feasible for large graphs. To circumvent this problem, a possible way is to approximate the filter $g_\theta$ by the Chebyshev polynomials up to the $K$-th order. Hence, Hammond *et al.* [36] proposed the following $K$-localized convolution on graphs:

$$g_\theta \otimes x = \sum_{k=0}^{K} \theta'_k T_k(L_{sym}) x \qquad (5)$$

where $T_k$ is the Chebyshev polynomials. Recently, Kipf and Welling [25] simplified Eq. (5) by limiting $K = 1$ and further approximating the largest eigenvalue $\lambda_{max} = 2$, where Eq. (4) can be rewritten as

$$g_\theta \otimes x = \theta \left(I + D^{-\frac{1}{2}}AD^{-\frac{1}{2}}\right) x \qquad (6)$$

Then, they consider a GCN with the following propagation rule

$$H^{(l+1)} = \sigma\left(D^{-\frac{1}{2}}AD^{-\frac{1}{2}}H^{(l)}W^{(l)}\right) \qquad (7)$$

where $H^{(l)}$ and $H^{(l+1)}$ are the input and output, respectively, and $W^{(l)}$ represents the weights.

With the rapid development of graph convolution, GCN has been widely applied to various application, such as recommender systems [37], semantic segmentation [34], etc. Besides, to our best knowledge, GCN has also been deployed for RSI semantic segmentation. However, Ma *et al.* [38] only utilized a fixed graph during the node convolution process, which led to the intrinsic relationship among the pixels cannot be precisely reflected. Moreover, the features extracted from the GCN are rich in semantic information, yet have blurred spatial and boundary details.

Generally speaking, CNN is used to process certain class of graph data with a fixed 2D raster structure though the adjacency matrix is not explicitly expressed, while GCN is used to process more generally unstructured graph data [39]. Similar to CNN, convolution kernels in GCN [40] work on all nodes of the graph, and the weights are shared in the calculation at each node, resulting in a remarkable reduction of the number of network parameters and effective avoidance of overfitting.

## III. METHODS

In order to solve the problems of blurred boundary and the inaccurate existence of thin and small scatter in RSIs segmentation, in this section, we propose a novel RSI-net for high-resolution RSIs semantic segmentation by incorporating both GCN and atrous CNN to extract both graph-level and contextual scene information from RSIs. As illustrated in Figure 2, the proposed RSI-net has two feature extraction encoder streams and a novel designed decoder part, i.e., GCN-based graph feature extractor and atrous CNN-based image feature extractor, in the encoder to learn the discriminative semantic information, while in the decoder the graph-level features generated by GCN-based encoder and the high/low-level image features produced by CNN-based encoder are decoded and fused to learn complementary representations for final semantic segmentation.

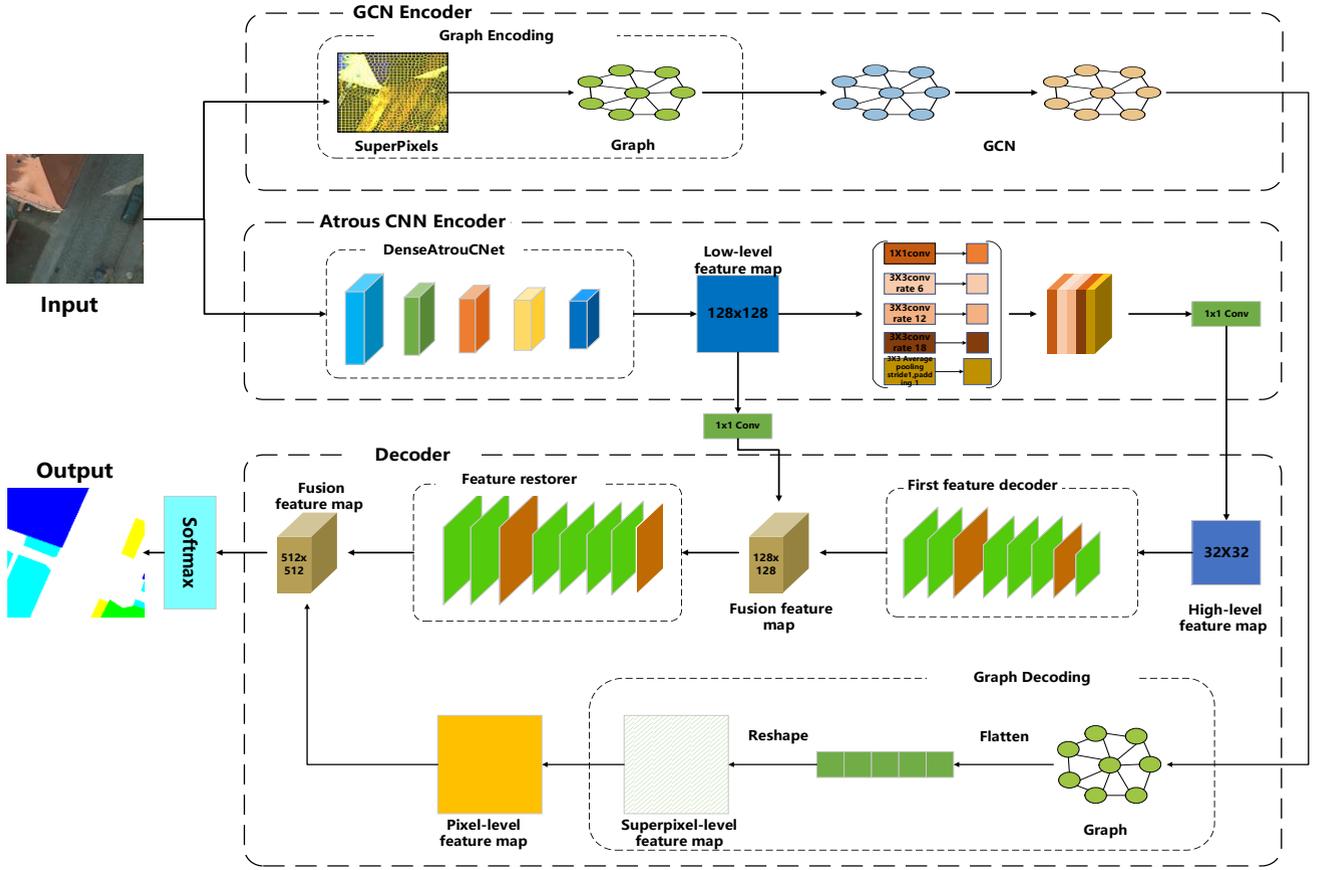

Fig. 2. Structure of the proposed RSI-Net. The network takes raw remote sensing images as inputs, and adopts two independently convolutional and graph convolutional streams as encoder to learn discriminative and complementary representations from multi-views. Then, the high-level feature produced by each stream are decoded and further fused for final pixel classification.

### A. GCN Encoder

The GCN encoder in the proposed RSI-Net aims to learn a latent graph representation of global context information in the scene image without relying on prior knowledge. In order to learn arbitrary graph features, the GCN encoder is designed to work on superpixel-based rather than the pixel-based nodes, for which simple linear iterative clustering (SLIC) [41] is utilized to partition RSIs into many spatially connected and spectrally similar structure-revealing superpixels. By establishing adjacency relations between superpixels, the image can be converted to an undirected graph $\mathcal{G} = (v, \varepsilon, A)$. Analogous to [28], let $S = \{S_i\}_{i=1}^{Z}$ denote the superpixel set and $V = [V_1, V_2, \cdots, V_Z]^T$. The corresponding node sets with the centroid of each superpixel are selected as the node. Moreover, in spired by [42, 43], a mapping association matrix $Q \in \mathbb{R}$ between pixels and superpixels is constructed with elements as

$$Q_{i,j} = \begin{cases} 1, & \text{if } \hat{X}_i \epsilon S_j \\ 0, & \text{otherwise} \end{cases} \quad \hat{X} = \text{Flatten}(X) \quad (8)$$

where Flatten(·) denotes the flattening operation of the RSI data, $\hat{X}_i$ the $i$th pixel in $\hat{X}$, $S_j$ the $j$th superpixel, and $Q_{i,j}$ the value of $Q$ at location $(i,j)$, respectively. Then, the node set $V$ can be calculated as

$$V = \text{Encode}(X, Q) = \hat{Q}^T \text{Flatten}(X) \quad (9)$$

where $\hat{Q}$ denotes the normalized $Q$ by column, and Encode(·, $Q$) the graph encoding used to translate RSI into the nodes of $\mathcal{G}$ according to $Q$. Then, we defined the adjacency mask $M \in \mathbb{R}$ as

$$M_{i,j} = \begin{cases} 1, & \text{if } S_i \text{ and } S_j \text{ are adjacent} \\ 0, & \text{otherwise} \end{cases} \quad (10)$$

where $M_{i,j}$ is the value of $M$ at location $(i,j)$. Note that $S_i$ and $S_i$ are adjacent when they share a common boundary. Moreover, we calculated the adjacency matrix as

$$A^{(l)} = \text{Sigmoid}\left((H^{(l-1)}W^{(l)})(H^{(l-1)}W^{(l)})^T\right) \odot M + I \quad (11)$$

where $H^{(l-1)}$ denotes the normalized input feature map of the $l$th graph convolutional layer, $W^{(l)}$ is the parameter to be learned, $M$ is the adjacency mask calculated by Eq. (10), the operator $\odot$ denotes Hadamard product, and $A^{(l)}$ is the learned adjacency matrix in the $l$th layer. The adjacency mask can cut off the association of nonadjacent nodes while keeping the connectivity between adjacent nodes. Finally, the final feature map is calculated according to Eq. (7).

Because our task is to predict pixels instead of superpixel-based nodes, the features of nodes should be assigned to pixels before segmentation. To propagate the node features to pixels, a graph decoding module is defined as

$$X^* = Decode(V, Q) \tag{12}$$

where $Decode(\cdot)$ denoted restoring operation of the spatial dimension of the flatten data. After the decoding procedure, the graph features can be projected back to the image space. The flow structure of the graph-based encoder was clearly shown in Figure 2.

### B. Atrous CNN Encoder

The atrous CNN encoder is based on the classic spatial pyramid pooling (ASPP) module [22] to expand receptive fields and extract more scaled image feature based semantic information. However, for complex background objects or small targets, the extracted features are neither dense nor detailed enough, resulting in insufficient segmentation. In view of this, in our proposed RSI-Net, improved DenseAtrousCNet is used instead of ASPP to perform down-sampling. DenseAtrousCNet utilizes a dense connection to cascade a series of atrous convolution and select a suitable atrous rate. By improving the perceptual field while reducing the problems arising from too sparse convolution kernels, DenseAtrousCNet can capture more local information to effectively mitigate the gap local information loss phenomenon.

In this stream of encoder, the feature map from DenseAtrousCNet is then fed into four densely connected atrous convolutions with one 1×1 kernel and three 3×3 kernels with atrous rates of 6, 12, and 18, respectively, as well as an image pooling, in parallel. The features from the above 5 channels are then aggregated and fed into convolution layer with a 3×3 kernel, a stride of 2 for downsamping, and a global average pooling to obtain a dense feature map, followed by a 1×1 convolution for dimensionality reduction. The detailed structure of the atrous CNN encoder is clearly descripted in Figure 2.

### C. Fused Decoder

Contextual information is known to be crucial for semantic segmentation of RSIs. Atrous convolution is an effective operation to expand receptive field and capture multi-scale contextual information. However, atrous convolution and GCN alone are not effective enough for RSIs, since they often generate large amount of small, thin and unclear misclassification objects. Therefore, we design a novel fused decoder based on the output features generated by atrous CNN and GCN from different perspectives to gradually restore the accurate RSI features. The decoder of the proposed RSI-Net is a fusion of the output of the two encoder streams. The decoder integrates both graph features and high-level as well as low-level image features, as shown in Figure 2 and detailed in Figure 3.

High-level image features from the CNN encoder stream form a feature map of 32×32 which is then fed into the first feature decoder block (showed in Figure 3 called Block 1) consisting of 10 deconvolution layers with kernels of increasing size, as detailed in Block 1 of Figure 3, resulting in a higher resolution feature map of 128×128 due to 4-fold upsampling. The first fusion operates on the deconvoluted feature map and the intermediate low-level feature map obtained by the DenseAtrousCNet in the CNN stream of the encoder, as shown in Figure 2. This fusion block, as shown in Figure 2 and the Block 2 of Figure 3, consists of a convolutional layer with a kernel of 1×1 and a stride of 1, which combines the two feature maps and performs dimensionality reduction, resulting in a feature map of 128×128. This feature map, representing fused low-level and high-level image features, is then fed into the feature restorer block, which restores the original image by 9 deconvolution layers. Specifically, as shown in Block 3 of Figure 3, the fused 128×128 feature map is first processed by a deconvolution layer with a 3×3 kernel and stride 2, then the upsampled 256×256 feature map is further processed by 4 deconvolution layers for refined image restoration, which is followed by another deconvolution layer of stride 2 to upsample the feature map to the original image size of 512×512, which is then further refined by 3 deconvolution layers, resulting in an image restoration from purely image features. At last, this image-feature based restoration is then fused with the graph-based restoration (from the GCN stream) in the second fusion block, as shown in Block 4 of Figure 3, and then classified with softmax loss function, resulting in final segmentation.

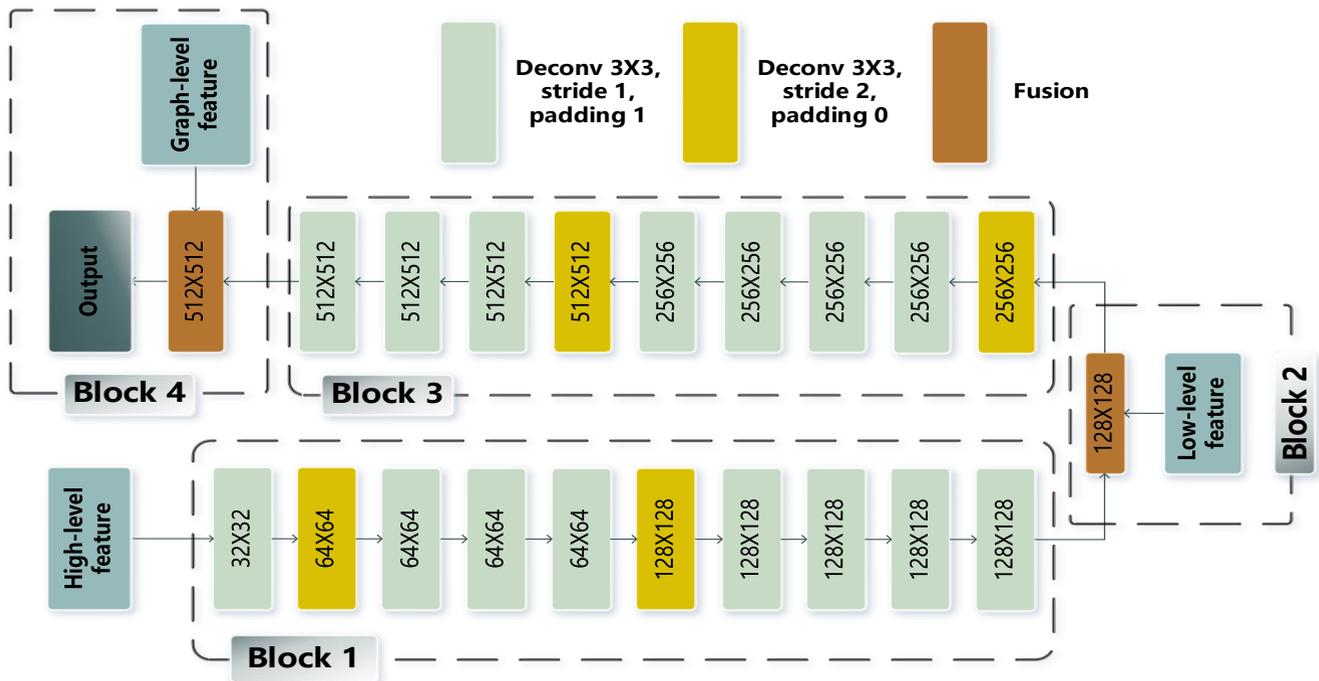

Fig. 3. Framework of the proposed decoder of RSI-Net. Low-level features are semantically enriched by embedding local focus from high image-level features and graph-level features.

## IV. EXPERIMENTAL RESULTS

### A. Datasets

To demonstrate the effectiveness of the proposed RSI-Net, experiments have been conducted on three public RSI datasets, i.e., the Potsdam Dataset, the Vaihingen Dataset, and the Gaofen Image Dataset, respectively.

#### 1) VAIHINGEN DATASET

The 2D semantic labeling benchmark Vaihingen dataset is released by the International Society for Photogrammetry and Remote Sensing (ISPRS, Vaihingen 2D Semantic Labeling Dataset) [44]. It includes high resolution true orthophoto tiles, digital surface models, and labeled ground truth. Each tile has three spectral bands: red (R), green (G), and near infrared (NIR). The spatial scale is approximately 2500 × 2000, with a ground sample distance (GSD) of 9 cm. The ground truth of labeled datasets is divided into six categories: impermeable surfaces, buildings, low vegetation, trees, cars, and clutter/background.

#### 2) POTSDAM DATASET

Potsdam data set (ISPRS, Potsdam 2D Semantic Labeling Dataset) [45] consists of 38 true orthophoto (TOP) tiles and the corresponding DSMs collected from a historic city with large building blocks. Each TOP image contains four spectral bands (red, green, blue, and near-infrared) and one DSM. All data files have the same spatial size, equal to 6000 × 6000 pixels. The ground sampling distance (GSD) of this data set is 5 cm. The ground truth is categorized as impermeable surfaces, buildings, low vegetation, trees, cars, and clutter/background.

#### 3) GAOFEN IMAGE DATASET

Gaofen Image Dataset (GID) is a large-scale land-cover dataset [46]. GID images have high intra-class diversity and low inter-class separability. The image size is 6800×7200 pixels. Multispectral provides images in the blue, green, red and near infrared bands. Since its launch in 2014, GF-2 has been used in various applications such as land survey, environmental monitoring, crop estimation, and construction planning. Five primary categories are annotated in the GID classification set: built-up, farmland, forest, meadow, and water, which are pixel-level labeled with five different colors: red, green, cyan, yellow, and blue, respectively.

### B. Implementation Details

In the experiments, we only use the IRRG three-channel color data and crop it to a standard image of size 512×512. In the GCN stream of the proposed RSI-Net, the segmentation scale of superpixels (shown in Figure 2 Graph Encoding block) is empirically fixed to 100. The number of graph convolutional layers is empirically set to 2, and the initial size of $W$ in Eq. (7) is 128×256, while detailed configurations of the DenseAtrousCNet are summarized in Figure 4. The specific parameters of atrous spatial pyramid pooling module, which extracts convolutional features at multiple scales by applying atrous convolution with different rates, are shown in Figure 2. Activation functions are all set to leaky rectified linear unit (Leaky ReLU), and Adam optimizer is used to train the network with the learning rate and number of training iterations empirically set to 0.0001 and 500, respectively, according to extensive experimental comparative analysis. The code is implemented on the PyTorch framework version 1.6.0 with Python 3.8, a GTX-1080Ti GPU under Ubuntu18.04, available at https://github.com/NZWANG/RSI-Net.

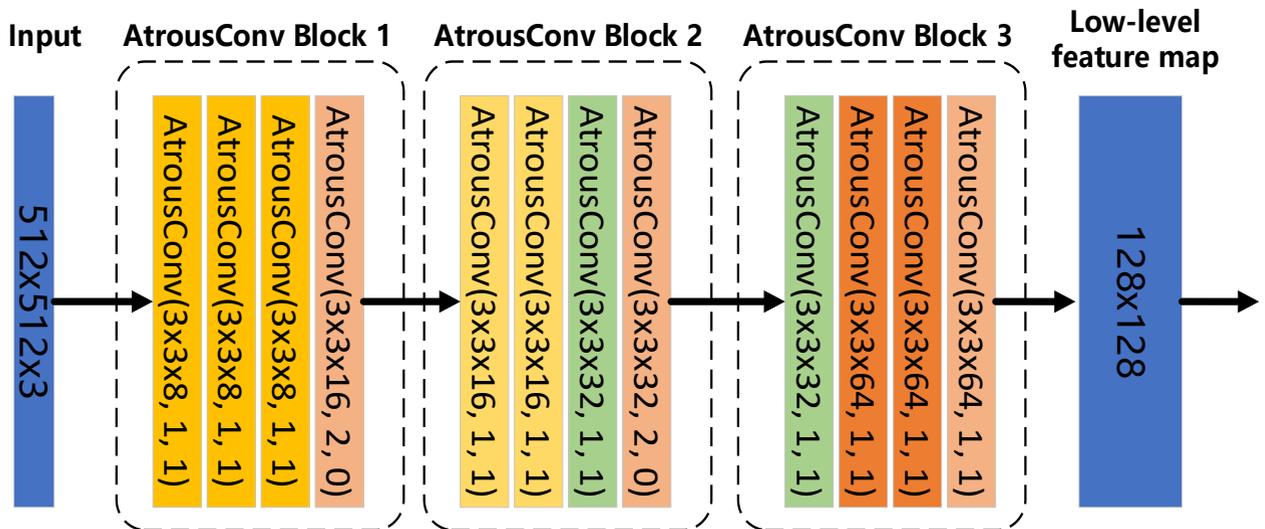

Fig. 4. Framework of the proposed DenseAtrousCNet of RSI-Net. AtrousConv (3 × 3 × $N$, $P$, $Q$) represents a kernel of 3 × 3, $N$ kernels, a stride of $P$, and an atrous rate of $Q$.

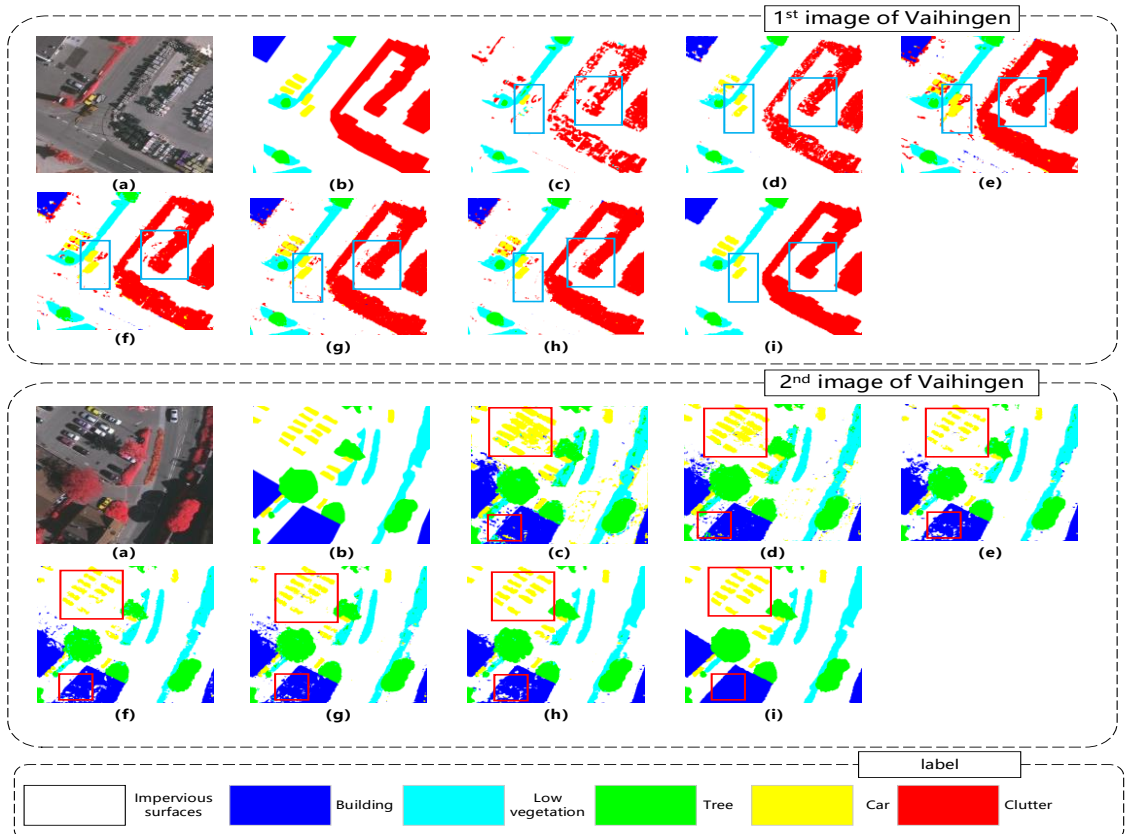

Fig. 5. Visualization of segmentation results from competing methods on the Vaihingen dataset. (a) Original image; (b) Ground truth; (c) FCN; (d) SegNet; (e) U-Net; (f) PSPNet; (g) DeepLab V3+; (h) GCN; (i) RSI-Net.

### C. Evaluation Metrics

Three widely-used evaluation metrics are applied to validate the performance of RSI-Net, i.e., per-class F1-score, overall accuracy (OA), F1 score, and Cohen's kappa coefficient ($\kappa$). The OA is calculated as

$$OA = \frac{TP+TN}{TP+FN+FP+TN} \quad (13)$$

where TP denotes the number of positive samples classified correctly, FN positive samples misclassified, FP negative samples misclassified, and TN negative samples classified correctly. The per-class F1 score is defined as the harmonic mean of precision and recall.

$$F1 = 2 \cdot \frac{precision \cdot recall}{precision+recall} \quad (14)$$

F1 score represents average F1 score of all classes, and a larger F1 score indicated the better performance of the segmentation algorithm. The kappa coefficient is calculated as

$$\kappa = \frac{p_o - p_e}{1 - p_e} \quad (15)$$

where $p_o$ is the relative agreement between the segmentation results and the real labels; $p_e$ is the hypothetical probability of a chance agreement. A kappa coefficient closer to 1 denotes the better agreement between the segmentation results and the ground truth. Note that F1 score, OA and $\kappa$ represent average scores of all classes.

### D. Competing Methods

The proposed RSI-Net is compared with several state-of-the-art RSI semantic segmentation methods including FCN, U-Net, SegNet, PSPNet, Deeplab v3+, and GCN, which are implemented by their original codes and trained on the datasets used in this paper. The overall classification

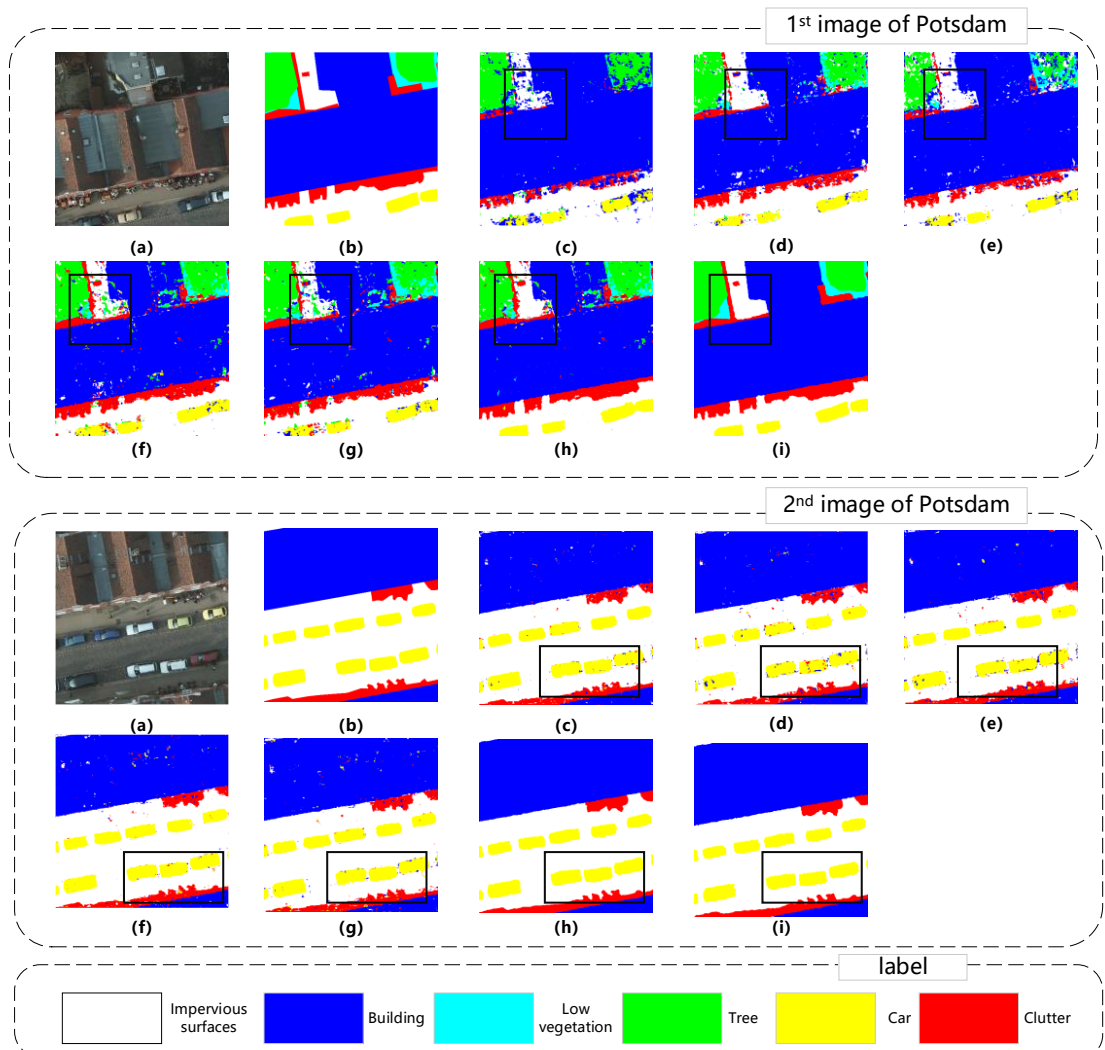

Fig. 6. Visualization of segmentation results from competing methods on the Potsdam dataset. (a) Original image; (b) Ground truth; (c) FCN; (d) SegNet; (e) U-Net; (f) PSPNet; (g) DeepLab V3+; (h) GCN; (i) RSI-Net.

performance (averaged over all classes) is quantitatively evaluated by the aforementioned metrics.

1) RESULTS ON THE VAIHINGEN DATASET

Quantitative results obtained by different methods on the Vaihingen dataset are summarized in Table 1, where the highest value in each row was highlighted in bold. It was quite evident that the results from RSI-Net are in exceptionally good agreement with expectation, with average OA, F1 score and κ of 91.83%, 90.30% and 89.46%, respectively, considerably higher than the competing methods under comparison, while the superior performance of RSI-Net is also well demonstrated by the per-class F1 scores.

Compared to FCN, SegNet, and U-Net, the OA of RSI-Net is dramatically increased by about 12.04%, 11.12% and 9.07%, respectively, and the per-class F1 score increased by at least 5%. PSPNet and DeepLab V3+ can enlarge the receptive field and capture more contextual information, yielding results better than FCN, SegNet, and U-Net but still not as good as RSI-Net. GCN operates on graph-structure data to obtain global and spatial information, resulting segmentation performance comparable to PSPNet and DeepLab V3+ but also inferior to RSI-Net.

To visualize segmentation from RSI-Net and the competing methods, results of two random samples from the Vaihingen dataset are presented in Figure 5, showing that contents in the two squares in (c)-(h) representing the six competing methods are considerably less consistent with the original images in (a) and ground truth segmentations in (b) when compared with RSI-Net in (i).

2) RESULTS ON THE POTSDAM DATASET

The same procedure is also performed on the Postdam dataset, with results presented in Table 2, showing that similar performance is obtained as in the Vaihingen dataset, *i.e*, RSI-Net is clearly superior in terms of the evaluation metrics including average OA, F1 score and κ, and also in most per-class F1 scores except the "low vegetation" class. PSPNet and DeepLab V3+ perform much better than FCN, SegNet and U-Net, GCN performs comparably to PSPNet and DeepLab V3+, but RSI-Net performs considerably better than all the competing methods.

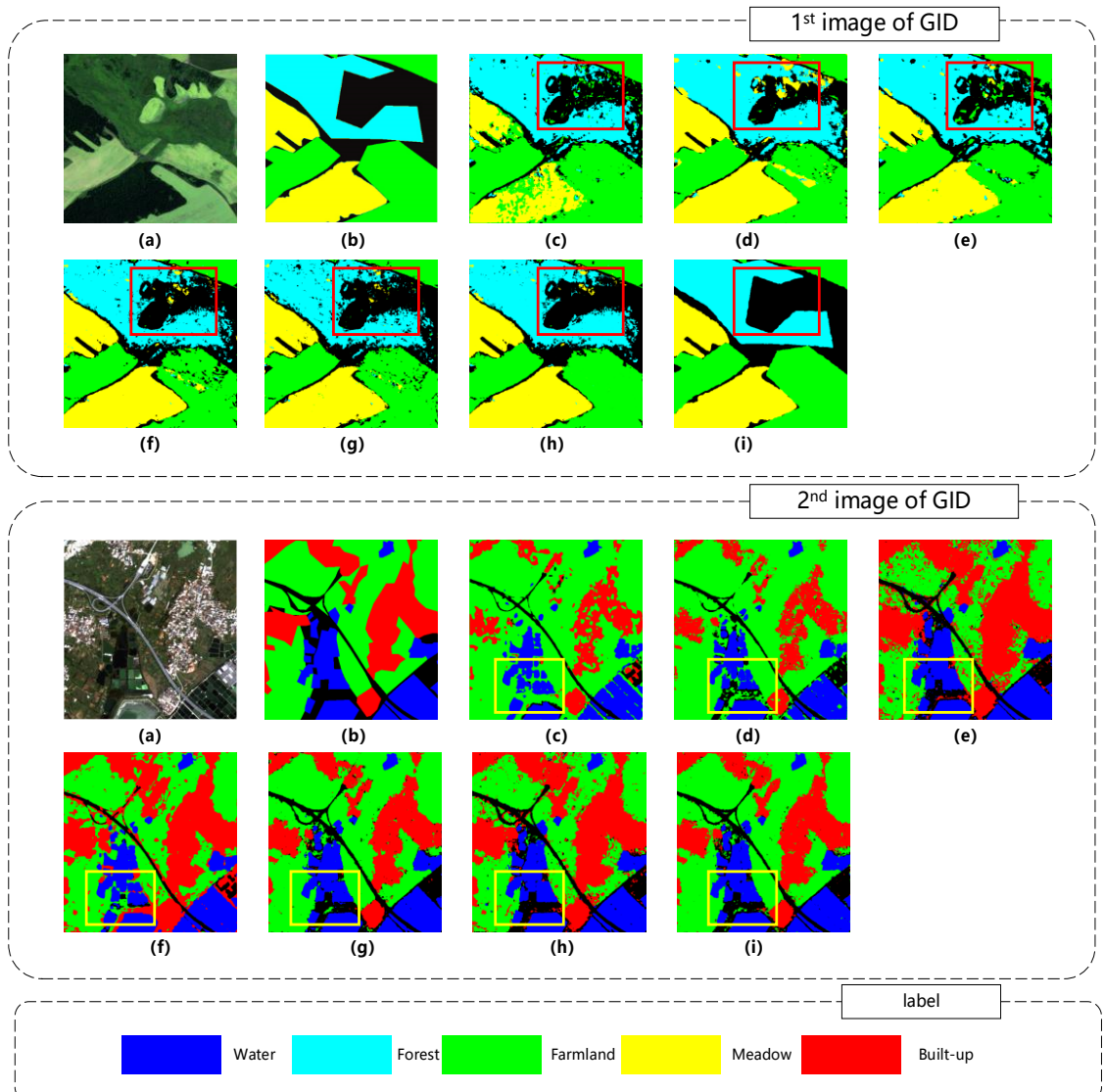

Fig. 7. Visualization of segmentation results from competing methods on the GID dataset. (a) Original image; (b) Ground truth; (c) FCN; (d) SegNet; (e) U-Net; (f) PSPNet; (g) DeepLab V3+; (h) GCN; (i) RSI-Net.

Visualization of segmentation results on two random images is demonstrated in Figure 6, showing that compared to the close to perfect segmentation of RSI-Net, CNN-based methods (FCN, SegNet, U-Net, PSPNet and DeepLab V3+) confuse cars and clutters with impervious surfaces, while GCN can separate cars from but still confuse cars with impervious surfaces.

### 3) RESULTS ON THE GID DATASET

The GID dataset, partially annotated with five classes, is used to evaluate the perfor-mance of RSI-Net for large-scale land-cover segmentation, which is presented in Table 3, demonstrating still considerable superiority to the competing results, in terms per-class F1 score as well as average κ, F1 score and OA.

As can be seen in the visualization in Figure 7, segmentations from FCN, SegNet and U-Net are relatively noisy and less consistent with the ground truth, for example more built-up areas are mislabeled as farmland. In the results of PSPNet and DeepLab V3+, some water and farmland areas located at the left bottom of the image are misclassified as built-up and non-image areas. Segmentations from GCN are also noisy in the farmland regions. Compared to these competing methods, RSI-Net provides segmentations much closer to the ground truth.

## VI. DISCUSSIONS

### A. SUPERPIXEL SIZE SELECTION in RSI-Net

The size of the superpixel has a potential impact on the resolution of the segmenta-tion performance, so in this subsection superpixel size of 100, 200, 300, 400, 500, and 600 are investigated, respectively, for its effect on segmentation performance of the proposed RSI-Net. F1 scores and kappa coefficients for the three datasets are plotted against different superpixel sizes in Figure 8, demonstrating that superpixel size has a potential effect on segmentation performance in RSI-Net, and superpixel size 100 seems work robustly on all three datasets in terms of both F1 score and kappa coefficient.

TABLE I
QUANTITATIVE RESULTS FROM COMPETING METHODS ON THE VAIHINGEN DATASET. PER-CLASS F1 SCORES ARE PRESENTED FOR THE SIX CLASSES. WHOLE DATASET OA, F1 SCORE, AND $\kappa$, AVERAGED OVER THE SIX CLASSES, ARE ALSO PRESENTED FOR RSI-NET AND THE SIX COMPETING METHODS. FOR EACH EVALUATION METRIC, THE HIGHEST VALUE IS HIGHLIGHTED IN BOLD.

| EVALUATION METRICS | FCN | SEGNET | U-NET | PSPNET | DEEPLAB V3+ | GCN | RSI-NET |
|---|---|---|---|---|---|---|---|
| F1 SCORE: IMPERVIOUS SURFACES | 92.15 | 94.66 | 93.74 | 91.44 | 94.35 | 91.37 | **98.03** |
| F1 SCORE: BUILDING | 89.86 | 91.34 | 90.66 | 94.11 | 94.34 | 94.38 | **95.52** |
| F1 SCORE: LOW VEGETATION | 85.72 | 83.76 | 86.12 | 91.52 | 91.40 | 91.28 | **92.79** |
| F1 SCORE: TREE | 78.47 | 78.66 | 79.46 | 87.50 | 87.91 | 87.84 | **92.81** |
| F1 SCORE: CAR | 66.05 | 68.94 | 70.01 | 71.76 | 76.81 | 78.02 | **83.46** |
| F1 SCORE: CLUTTER | 68.79 | 70.75 | 72.33 | 73.04 | 75.26 | 74.39 | **79.16** |
| OA (%) | 79.79 | 80.71 | 82.76 | 88.91 | 89.34 | 88.94 | **91.83** |
| AVERAGE F1 SCORE (%) | 80.17 | 81.35 | 82.05 | 84.89 | 86.67 | 86.21 | **90.30** |
| $\kappa$ (%) | 74.57 | 75.52 | 76.43 | 83.85 | 85.97 | 85.75 | **89.46** |

TABLE II
QUANTITATIVE RESULTS FROM COMPETING METHODS ON THE POTSDAM DATASET. PER-CLASS F1 SCORES ARE PRESENTED FOR THE SIX CLASSES. WHOLE DATASET OA, F1 SCORE, AND $\kappa$, AVERAGED OVER THE SIX CLASSES, ARE ALSO PRESENTED FOR RSI-NET AND THE SIX COMPETING METHODS. FOR EACH EVALUATION METRIC, THE HIGHEST VALUE IS HIGHLIGHTED IN BOLD.

| EVALUATION METRICS | FCN | SEGNET | U-NET | PSPNET | DEEPLAB V3+ | GCN | RSI-NET |
|---|---|---|---|---|---|---|---|
| F1 SCORE: IMPERVIOUS SURFACES | 93.21 | 94.46 | 94.21 | 92.37 | 95.43 | 92.77 | **99.13** |
| F1 SCORE: BUILDING | 88.63 | 90.34 | 92.37 | 95.19 | 93.89 | 94.84 | **96.58** |
| F1 SCORE: LOW VEGETATION | 86.72 | 85.64 | 87.22 | 92.73 | 93.83 | **93.86** | 93.79 |
| F1 SCORE: TREE | 77.87 | 76.37 | 80.18 | 88.13 | 89.97 | 89.79 | **93.81** |
| F1 SCORE: CAR | 68.53 | 69.48 | 69.01 | 77.17 | 78.81 | 80.87 | **84.44** |
| F1 SCORE: CLUTTER | 69.78 | 72.58 | 72.17 | 75.43 | 77.66 | 79.47 | **81.23** |
| OA (%) | 80.13 | 82.11 | 83.68 | 87.99 | 90.44 | 91.43 | **93.31** |
| AVERAGE F1 SCORE (%) | 80.79 | 81.48 | 82.52 | 86.83 | 88.26 | 88.60 | **91.49** |
| $\kappa$ (%) | 75.71 | 77.73 | 78.39 | 85.38 | 87.77 | 88.54 | **90.96** |

TABLE III
QUANTITATIVE RESULTS FROM COMPETING METHODS ON THE GID DATASET. PER-CLASS F1 SCORES ARE PRESENTED FOR THE SIX CLASSES. WHOLE DATASET OA, F1 SCORE, AND $\kappa$, AVERAGED OVER THE SIX CLASSES, ARE ALSO PRESENTED FOR RSI-NET AND THE SIX COMPETING METHODS. FOR EACH EVALUATION METRIC, THE HIGHEST VALUE IS HIGHLIGHTED IN BOLD.

| EVALUATION METRICS | FCN | SEGNET | U-NET | PSPNET | DEEPLAB V3+ | GCN | RSI-NET |
|---|---|---|---|---|---|---|---|
| F1 SCORE: BUILT-UP | 71.11 | 72.34 | 73.21 | 80.03 | 85.02 | 86.31 | **87.19** |
| F1 SCORE: FARMLAND | 86.89 | 85.46 | 87.39 | 90.74 | 91.91 | 92.14 | **93.04** |
| F1 SCORE: FOREST | 70.16 | 76.01 | 77.03 | 80.63 | 82.42 | 83.36 | **85.33** |
| F1 SCORE: MEADOW | 71.55 | 70.63 | 70.91 | 81.88 | 83.36 | 84.46 | **86.06** |
| F1 SCORE: WATER | 88.37 | 87.67 | 88.04 | 90.11 | 93.02 | 91.07 | **95.15** |
| OA (%) | 83.31 | 85.04 | 86.11 | 88.37 | 90.71 | 91.46 | **93.67** |
| AVERAGE F1 SCORE (%) | 77.62 | 78.42 | 79.32 | 84.68 | 87.15 | 87.47 | **89.35** |
| $\kappa$ (%) | 75.37 | 76.33 | 76.46 | 83.43 | 86.31 | 87.37 | **90.37** |

Since larger superpixels lead to smaller graphs which preserve bigger objects and less noise, while smaller superpixels preserve smaller objects and more noise, generally the performance on the Vaihingen and Potsdam datasets decreases with the increase of superpixel size due to the fact that many small objects are existed in the datasets. In contrast, performance on the GID dataset roughly increases with the increase of superpixel size, in accordance with the large proportion of big objects in GID. To prevent the RSI-Net from generating over-smoothed segmentation maps, the superpixel size is empirically set to 100 in all experiments in this study, which provides good results for the Vaihigen and Potsdam datasets and acceptable performance for the GID dataset.

## B. Time Cost Comparison

Though PSPNet and DeepLab V3+ perform comparably to GCN on all three datasets, GCN is computationally faster while PSPNet and DeepLab V3+ are often criticized for their high usage of both computation power of memory. Our proposed RSI-Net integrates both the advantage of GCN using graph-level features and the advantage of PSPNet and DeepLab V3+ of expanded receptive fields to obtain further improved segmentation while still computationally efficient, as demonstrated in Table 4, where running time of RSI-Net as well as the state-of-the-art competing methods are presented for all the three datasets, showing that RSI-Net runs slower than GCN but much faster than all other competing methods.

## C. Effects of Different Components of RSI-Net

In order to explore each stream's affecting performance of RSI-Net, we designed four variations. Note that we only analyzed the effects of each component so that the usage of other parameters was referred to previous work. Table V illustrated the OA, average F1-score and $\kappa$ of RSI-Net based on three datasets.

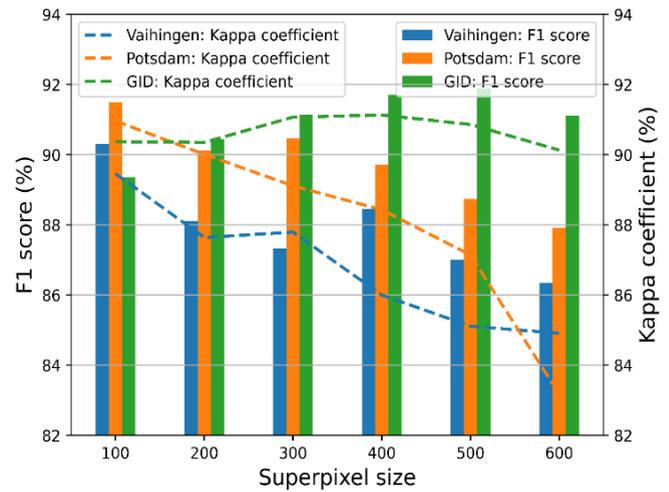

Fig. 8. The F1 scores and kappa coefficients of RSI-Net with different superpixel' sizes on three public datasets.

TABLE IV
COMPARISON OF THE RUNNING TIME (IN SECONDS) OF RSI-NET AND THE COMPETING METHODS ON THE THREE PUBLIC DATASETS.

| DATASETS \ METHODS | FCN | SEGNET | U-NET | PSPNET | DEEPLAB V3+ | GCN | RSI-NET |
|---|---|---|---|---|---|---|---|
| VAIHINGEN | 2137.2 | 3680.0 | 4219.0 | 3137.2 | 4369.8 | 672.3 | 1288.7 |
| POTSDAM | 2913.7 | 4003.2 | 4831.3 | 3733.8 | 4937.4 | 814.5 | 1672.0 |
| GID | 3475.3 | 4476.3 | 5317.4 | 4261.3 | 5341.8 | 883.2 | 1698.1 |

As listed in Table V, without GCN components, Atrous CNN components or DenseAtrousCNet components, RSI-Net, precisely, boiled down to a very unstable level, and the OA, average F1 score and $\kappa$ relied heavily on the CNN extractors, i.e., the OA, F1-Score and $\kappa$ are decreased by at least 5% without DenseAtrousCNet components, respectively. In addition, without GCN components or Atrous CNN components, the OA, F1-score and $\kappa$ of RSI-Net were all affected to a greater or lesser extent. Moreover, GID dataset was more sensitive to GCN components, whereas Vaihingen datasets was more stable. For instance, RSI-Net obtained an OA of 93.67% for the GID dataset decreased by 6.64% without GCN components, whereas the OA obtained by the Vaihingen dataset was decreased by only 1.7%. Accordingly, the combination of the all components could be more superior over other variants, and it was more stable on each dataset.

## VII. CONCLUSION

Striven to enhance the distinguishability of the correlation between adjacent land covers and to deal with the boundary blur in high-resolution remote sensing images, a novel RSI-Net is proposed and implemented in this paper. First of all, inspired by the ad-vantages of GCN, the graph-based encoder is designed as one key stream to model the pixel-superpixel dependencies to construct the graph, and propagate the contextual in-formation, which effectively produces the enhanced superpixel representations. Further, taking advantage of CNN, the DenseAtrousCNet and atrous convolution are designed as the other stream to deal with large variations in target scales in the semantic segmentation of remote sensing images. Moreover, a novel decoder combining both image-level and graph-level information is designed to boost the efficiency of semantic segmentation, where deconvolution operations are embedded to recover feature maps to the original spatial resolution. Finally, comparisons with several state-of-the-art RSI semantic segmentation methods are performed on three public RSI datasets to demonstrate the superior performance of RSI-Net on semantic segmentation of high-resolution remote sensing images.

TABLE V
CLASSIFICATION RESULTS WITHOUT VARIOUS COMPONENTS ON EACH DATASET.

| Datasets | | Without GCN | Without Atrous CNN | Without DenseAtrousCNet | RSI-Net |
|---|---|---|---|---|---|
| Vaihingen | OA (%) | 90.13 | 89.22 | 85.47 | **91.83** |
| | F1 Score (%) | 86.02 | 85.71 | 83.24 | **90.30** |
| | $\kappa \times 100$ | 86.17 | 84.38 | 83.01 | **89.46** |
| Potsdam | OA (%) | 89.41 | 91.06 | 85.02 | **93.31** |
| | F1 Score (%) | 89.73 | 88.70 | 84.28 | **91.49** |
| | $\kappa \times 100$ | 87.31 | 88.05 | 85.58 | **90.96** |
| GID | OA (%) | 87.03 | 90.17 | 86.23 | **93.67** |
| | F1 Score (%) | 85.46 | 88.33 | 84.82 | **89.35** |
| | $\kappa \times 100$ | 86.17 | 88.31 | 83.68 | **90.37** |

According to vast experimental findings, contextual information is quite essential to enhance the segmentation in pixel-wise semantics. Therefore, the pixel-superpixel correlation encoding model of feature maps and its decoding model should be explored in the future. Besides, a more advanced feature map fusion decoder can potentially further improve the performance of RSI-Net. Finally, it is necessary to utilize pruning and knowledge distillation methods to simplify our model in future.

**ACKNOWLEDGMENT**
We appreciate the editor's and reviewers' constructive comments, which improve the quality of this article. We would like to thank Prof. Gou Jianping for his advice and help.

**REFERENCES**
[1] M. Wu, C. Zhang, J. Liu, L. Zhou, and X. Li, "Towards accurate high resolution satellite image semantic segmentation," *IEEE Access,* vol. 7, pp. 55609-55619, 2019.
[2] M. Wurm, T. Stark, X. X. Zhu, M. Weigand, and H. Taubenböck, "Semantic segmentation of slums in satellite images using transfer learning on fully convolutional neural networks," *ISPRS journal of photogrammetry and remote sensing,* vol. 150, pp. 59-69, 2019.
[3] H. Im and H. Yang, "Analysis and Optimization of CNN-based Semantic Segmentation of Satellite Images," in *2019 International Conference on Information and Communication Technology Convergence (ICTC)*, 2019: IEEE, pp. 218-220.
[4] S. Chantharaj *et al.*, "Semantic segmentation on medium-resolution satellite images using deep convolutional networks with remote sensing derived indices," in *2018 15th International joint conference on computer science and software engineering (JCSSE)*, 2018: IEEE, pp. 1-6.
[5] D. Marcos, M. Volpi, B. Kellenberger, and D. Tuia, "Land cover mapping at very high resolution with rotation equivariant CNNs: Towards small yet accurate models," *ISPRS journal of photogrammetry and remote sensing,* vol. 145, pp. 96-107, 2018.
[6] X. Liu *et al.*, "Classifying urban land use by integrating remote sensing and social media data," *International Journal of Geographical Information Science,* vol. 31, no. 8, pp. 1675-1696, 2017.
[7] O. Tasar, Y. Tarabalka, and P. Alliez, "Incremental learning for semantic segmentation of large-scale remote sensing data," *IEEE Journal of Selected Topics in Applied Earth Observations and Remote Sensing,* vol. 12, no. 9, pp. 3524-3537, 2019.
[8] G. N. Kouziokas and K. Perakis, "Decision support system based on artificial intelligence, GIS and remote sensing for sustainable public and judicial management," *European Journal of Sustainable Development,* vol. 6, no. 3, pp. 397-397, 2017.
[9] S. M. Azimi, P. Fischer, M. Körner, and P. Reinartz, "Aerial LaneNet: Lane-marking semantic segmentation in aerial imagery using wavelet-enhanced cost-sensitive symmetric fully convolutional neural networks," *IEEE Transactions on Geoscience and Remote Sensing,* vol. 57, no. 5, pp. 2920-2938, 2018.
[10] Z. Chen, C. Wang, J. Li, N. Xie, Y. Han, and J. Du, "Reconstruction Bias U-Net for Road Extraction From Optical Remote Sensing Images," *IEEE Journal of Selected Topics in Applied Earth Observations and Remote Sensing,* vol. 14, pp. 2284-2294, 2021.
[11] Y. Wei, K. Zhang, and S. Ji, "Simultaneous road surface and centerline extraction from large-scale remote sensing images using CNN-based segmentation and tracing," *IEEE Transactions on Geoscience and Remote Sensing,* vol. 58, no. 12, pp. 8919-8931, 2020.
[12] C. Yoo, D. Han, J. Im, and B. Bechtel, "Comparison between convolutional neural networks and random forest for local climate zone classification in mega urban areas using Landsat images," *ISPRS Journal of Photogrammetry and Remote Sensing,* vol. 157, pp. 155-170, 2019.
[13] Y. Wang, W. Yu, and Z. Fang, "Multiple kernel-based SVM classification of hyperspectral images by combining spectral, spatial, and semantic information," *Remote Sensing,* vol. 12, no. 1, p. 120, 2020.
[14] C. Zheng, Y. Zhang, and L. Wang, "Multigranularity multiclass-layer markov random field model for semantic segmentation of remote sensing images," *IEEE Transactions on Geoscience and Remote Sensing,* vol. 59, no. 12, pp. 10555-10574, 2020.
[15] Y. Kong, B. Zhang, B. Yan, Y. Liu, H. Leung, and X. Peng, "Affiliated Fusion Conditional Random Field for Urban UAV Image Semantic Segmentation," *Sensors,* vol. 20, no. 4, p. 993, 2020.
[16] Y. LeCun, L. Bottou, Y. Bengio, and P. Haffner, "Gradient-based learning applied to document recognition," *Proceedings of the IEEE,* vol. 86, no. 11, pp. 2278-2324, 1998.
[17] A. Krizhevsky, I. Sutskever, and G. E. Hinton, "Imagenet classification with deep convolutional neural networks," *Advances in neural information processing systems,* vol. 25, pp. 1097-1105, 2012.
[18] J. Long, E. Shelhamer, and T. Darrell, "Fully convolutional networks for semantic segmentation," in *Proceedings of the IEEE conference on computer vision and pattern recognition*, 2015, pp. 3431-3440.
[19] V. Badrinarayanan, A. Kendall, and R. Cipolla, "Segnet: A deep convolutional encoder-decoder architecture for image segmentation," *IEEE transactions on pattern analysis and machine intelligence,* vol. 39, no. 12, pp. 2481-2495, 2017.
[20] O. Ronneberger, P. Fischer, and T. Brox, "U-net: Convolutional networks for biomedical image segmentation," in *International Conference on Medical image computing and computer-assisted intervention*, 2015: Springer, pp. 234-241.
[21] L.-C. Chen, G. Papandreou, I. Kokkinos, K. Murphy, and A. L. Yuille, "Semantic image segmentation with deep convolutional nets and fully connected crfs," *arXiv preprint arXiv:1412.7062,* 2014. [Online]. Available:https://arxiv.org/abs/1412.7062.
[22] L.-C. Chen, G. Papandreou, I. Kokkinos, K. Murphy, and A. L. Yuille, "Deeplab: Semantic image segmentation with deep convolutional nets, atrous convolution, and fully connected crfs," *IEEE transactions on pattern analysis and machine intelligence,* vol. 40, no. 4, pp. 834-848, 2017.
[23] L.-C. Chen, G. Papandreou, F. Schroff, and H. Adam, "Rethinking atrous convolution for semantic image segmentation," *arXiv preprint arXiv:1706.05587,* 2017. [Online]. Available: https://arxiv.org/abs/1706.05587.
[24] S. Zheng *et al.*, "Conditional random fields as recurrent neural networks," in *Proceedings of the IEEE international conference on computer vision*, 2015, pp. 1529-1537.
[25] T. N. Kipf and M. Welling, "Semi-supervised classification with graph convolutional networks," *arXiv preprint arXiv:1609.02907,* 2016. [Online]. Available: https://arxiv.org/abs/1609.02907.
[26] F. Wu, A. Souza, T. Zhang, C. Fifty, T. Yu, and K. Weinberger, "Simplifying graph convolutional networks," in *International conference on machine learning*, 2019: PMLR, pp. 6861-6871.
[27] L. Mou, X. Lu, X. Li, and X. X. Zhu, "Nonlocal graph convolutional networks for hyperspectral image classification," *IEEE Transactions on Geoscience and Remote Sensing,* vol. 58, no. 12, pp. 8246-8257, 2020.
[28] S. Wan, C. Gong, P. Zhong, B. Du, L. Zhang, and J. Yang, "Multiscale dynamic graph convolutional network for hyperspectral image

classification," *IEEE Transactions on Geoscience and Remote Sensing,* vol. 58, no. 5, pp. 3162-3177, 2019.

[29] H. Zhao, J. Shi, X. Qi, X. Wang, and J. Jia, "Pyramid scene parsing network," in *Proceedings of the IEEE conference on computer vision and pattern recognition*, 2017, pp. 2881-2890.
[30] F. Yu, V. Koltun, and T. Funkhouser, "Dilated residual networks," in *Proceedings of the IEEE conference on computer vision and pattern recognition*, 2017, pp. 472-480.
[31] P. Wang *et al.*, "Understanding convolution for semantic segmentation," in *2018 IEEE winter conference on applications of computer vision (WACV)*, 2018: IEEE, pp. 1451-1460.
[32] N. Tremblay, P. Gonçalves, and P. Borgnat, "Design of graph filters and filterbanks," in *Cooperative and Graph Signal Processing*: Elsevier, 2018, pp. 299-324.
[33] D. I. Shuman, S. K. Narang, P. Frossard, A. Ortega, and P. Vanderghenyst, "The emerging filed of signal processing on graphs," *IEEE Signal Processing Magazine,* 2013.
[34] J. Bruna, W. Zaremba, A. Szlam, and Y. LeCun, "Spectral networks and locally connected networks on graphs," *arXiv preprint arXiv:1312.6203,* 2013.
[35] A. Sandryhaila and J. M. Moura, "Discrete signal processing on graphs: Graph filters," in *2013 IEEE International Conference on Acoustics, Speech and Signal Processing*, 2013: IEEE, pp. 6163-6166.
[36] D. K. Hammond, P. Vandergheynst, and R. Gribonval, "Wavelets on graphs via spectral graph theory," *Applied and Computational Harmonic Analysis,* vol. 30, no. 2, pp. 129-150, 2011.
[37] F. Scarselli, M. Gori, A. C. Tsoi, M. Hagenbuchner, and G. Monfardini, "Computational capabilities of graph neural networks," *IEEE Transactions on Neural Networks,* vol. 20, no. 1, pp. 81-102, 2008.
[38] F. Ma, F. Gao, J. Sun, H. Zhou, and A. Hussain, "Attention graph convolution network for image segmentation in big SAR imagery data," *Remote Sensing,* vol. 11, no. 21, p. 2586, 2019.
[39] Z. Wu, S. Pan, F. Chen, G. Long, C. Zhang, and S. Y. Philip, "A comprehensive survey on graph neural networks," *IEEE transactions on neural networks and learning systems,* vol. 32, no. 1, pp. 4-24, 2020.
[40] K. Xu, W. Hu, J. Leskovec, and S. Jegelka, "How powerful are graph neural networks?," *arXiv preprint arXiv:1810.00826,* 2018. [Online]. Available: https://arxiv.org/abs/1810.00826.
[41] R. Achanta, A. Shaji, K. Smith, A. Lucchi, P. Fua, and S. Süsstrunk, "SLIC superpixels compared to state-of-the-art superpixel methods," *IEEE transactions on pattern analysis and machine intelligence,* vol. 34, no. 11, pp. 2274-2282, 2012.
[42] V. Jampani, D. Sun, M.-Y. Liu, M.-H. Yang, and J. Kautz, "Superpixel sampling networks," in *Proceedings of the European Conference on Computer Vision (ECCV)*, 2018, pp. 352-368.
[43] Q. Liu, L. Xiao, J. Yang, and Z. Wei, "CNN-enhanced graph convolutional network with pixel-and superpixel-level feature fusion for hyperspectral image classification," *IEEE Transactions on Geoscience and Remote Sensing,* 2020.
[44] I. Vaihingen, "2D semantic labeling dataset," ed: Accessed: Apr, 2018.
[45] I. Potsdam, "2D Semantic Labeling Dataset," ed: Accessed: Apr, 2018.
[46] X.-Y. Tong *et al.*, "Land-cover classification with high-resolution remote sensing images using transferable deep models," *Remote Sensing of Environment,* vol. 237, p. 111322, 2020.

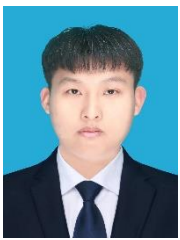

**SHUANG HE** received the bachelor's degree in Information and Computing Science from the Jiangsu Ocean University, Lianyungang, China, in 2018. He is currently pursuing the master's degree in Surveying and Remote Sensing Engineering with Jiangsu Ocean University.

His research interests include deep learning, remote sensing image processing, coastal wetland ecological environment monitoring and evaluation.

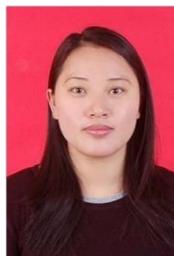

**XIA LU** received the B.S. degree in Exploration Engineering from Jilin University, Jilin, China, in 1997, the M.S. degree in Cartography and Geographic Information Engineering from the China University of Geoscience (Beijing), Beijing, China, in 2004, and the Ph.D. degree in Photogrammetry and Remote Sensing from China University of Mining (Beijing), Beijing, China, in 2008.

She is currently a Professor with School of Marine Technology and Geomatics, Jiangsu Ocean University. Her research interests include three-dimensional monitoring and evaluation of marine environment, quantitative remote sensing of coastal wetland ecology.

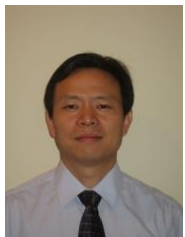

**JASON GU** is currently a Professor of robotics and assistive technology with the Department of Electrical and Computer Engineering, Dalhousie University. He has published more than 300 journals, book chapters and conference articles. His research interests include biomedical engineering, biosignal processing, rehabilitation engineering, neural networks, robotics, mechatronics, and control. He is the IEEE member of SMC and RAS. He is also a Fellow of the Engineering Institute of Canada and the Canada Academy of Engineering. He has been an Editor of the Journal of Control and Intelligent Systems, the Transactions on CSME, the IEEE TRANSACTIONS ON MECHATRONICS, the IEEE SMC Magazine, the International Journal of Robotics and Automation, and IEEE ACCESS. He is also the IEEE Canada President-Elect, from 2018 to 2019.

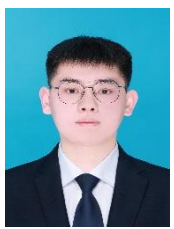

**HAITONG TANG** received the bachelor's degree in communication engineering from the Anhui Sanlian University, Hefei, China, in 2018. He is currently pursuing the master's degree in Surveying and Remote Sensing Engineering with Jiangsu Ocean University.

His research interests include deep learning, computer vision and convolutional sparse coding.

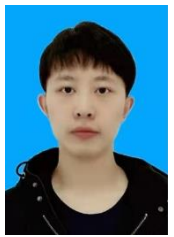

**QIN YU** received the bachelor's degree in electronic information engineering from the Huaiyin Normal University, in 2019. He is currently pursuing the master's degree in pattern recognition and intelligent systems with Jiangsu Ocean University.

His research interests include computational neuroscience, time-frequency analysis and artificial intelligence.

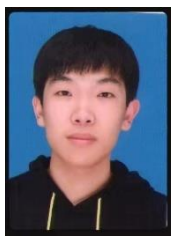

**KAIYUE LIU** graduated from Shanxi University of Engineering and Technology in 2018 with a bachelor's degree in Surveying and Mapping Engineering. He is currently studying for a master's degree in Civil Engineering at Jiangsu Ocean University. His research interests include the field of remote sensing, machine learning, object detection and ROS systems.

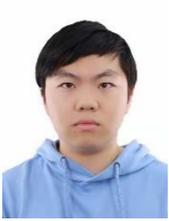

**HAOZHOU DING** is an undergraduate from Nanjing University of Information Science and Technology, majoring in Software Engineering. His research interests include remote sensing, machine learning, and computer vision.

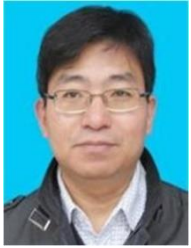

**CHUNQI CHANG** received the B.Sc. and M.Sc. degrees from the Department of Electronic Engineering and Information Science, University of Science and Technology of China, in 1992 and 1995, respectively, and the Ph.D. degree from The University of Hong Kong, in 2001. He is currently a Professor and Director of Laboratory of Neuroinformatics and Neuroengineering, School of Biomedical Engineering, Shenzhen University.

His research interests include neuroinformatics, neural engineering, and biomedical signal processing.

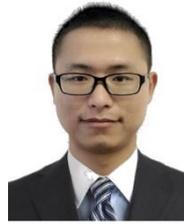

**NIZHUAN WANG** received the B.S. degree in computer science and technology from Heilongjiang University, Harbin, China, in 2010. He received the M.S. degree in computer application technology from Shanghai Maritime University, Shanghai, China, in 2012, where he also received the Ph.D. degree in communications, information engineering, and control, in January 2016. From 2022, he is currently a Research Associate Professor, School of Biomedical Engineering, ShanghaiTech University. Before that, he was an assistant professor of Shenzhen University and professor of Jiangsu Ocean University, respectively.

His research interests include neuroimaging, hyperspectral imaging, BCI and machine learning. He has published over 40 scientific papers on many journals, i.e., Human Brain Mapping, IEEE Journal of Biomedical and Health Informatics, IEEE Transactions on Biomedical Engineering, Journal of Neuroscience Methods, Magnetic Resonance Imaging, etc.